\documentclass[sigconf]{acmart}

\setcopyright{none}
\renewcommand\footnotetextcopyrightpermission[1]{} 
\makeatletter
\def\@copyrightspace{\relax}
\makeatother

\begin{document}

\title{Transfer Learning across Several Centuries: Machine and Historian Integrated Method to Decipher Royal Secretary’s Diary}

\author{Sojung Lucia Kim}
\authornote{corresponding author}
\email{sojung.kim@snu.ac.kr}
\orcid{0000-0002-5528-101X}
\affiliation{%
  \institution{Seoul National University}
  \institution{Nara.Lab}    
  \streetaddress{Gwanak-ro, Gwanak-gu 1}
  \city{Seoul}  
  \country{South Korea}
  \postcode{08826}
}

\author{Taehong Jang}
\authornote{corresponding author}
\affiliation{%
  \institution{Nara.Lab}    
  \streetaddress{Bongensa-ro 68gil23}
  \city{Seoul}  
  \country{South Korea}}
  \email{starbirdnara@gmail.com}

\author{Joonmo Ahn}
\affiliation{%
	\institution{Nara.Lab}    
	\streetaddress{Bongensa-ro 68gil23}
	\city{Seoul}  
	\country{South Korea}}
\email{jmahn.nara@gmail.com}

\author{Hyungil Lee}
\affiliation{%
	\institution{Nara.Lab}    
	\streetaddress{Bongensa-ro 68gil23}
	\city{Seoul}  
	\country{South Korea}}
\email{hyungil1004@gmail.com}

\author{Jaehyuk Lee}
\affiliation{%
	\institution{Nara.Lab}    
	\streetaddress{Bongensa-ro 68gil23}
	\city{Seoul}  
	\country{South Korea}}
\email{jaehyuklee2go@gmail.com}

\renewcommand{\shortauthors}{Kim et al.}

\begin{abstract}
  A named entity recognition and classification plays the first and foremost important role in capturing semantics in data and anchoring in translation as well as downstream study for history. However, NER in historical text has faced challenges such as scarcity of annotated corpus, multilanguage variety, various noise, and different convention far different from the contemporary language model. This paper introduces Korean historical corpus (Diary of Royal secretary which is named SeungJeongWon) recorded over several centuries and recently added with named entity information as well as phrase markers which historians carefully annotated. We fined-tuned the language model on history corpus, conducted extensive comparative experiments using our language model and pretrained muti-language models. We set up the hypothesis of combination of time and annotation information and tested it based on statistical t test. Our finding shows that phrase markers clearly improve the performance of NER model in predicting unseen entity in documents written far different time period. It also shows that each of phrase marker and corpus-specific trained  model does not improve the performance. We discuss the future research directions and practical strategies to decipher the history document. 
\end{abstract}

\ccsdesc[500]{Computing methodologies~Machine learning}
\ccsdesc[500]{Information systems~Specialized information retrieval}

\keywords{history NER, transfer learning, comparative experiment}
\begin{teaserfigure}
	\includegraphics[width=\textwidth]{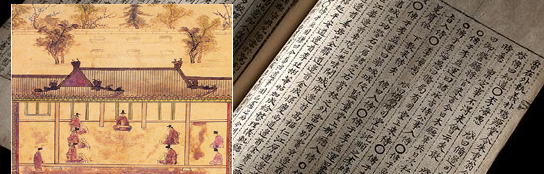}
\caption{Royal secretary(left) and their diary(right)} \label{fig1}
  \Description{secretary and their diary}
  \label{fig:teaser}
\end{teaserfigure}



\settopmatter{printfolios=true}
\pagestyle{plain}
\maketitle

\section{Introduction}
Named Entity Recognition (NER) is a natural language processing (NLP) technique that involves the identification and classification of named entities within text data, such as names of people, organizations, locations, and other entities of interest. 
Thus it is considered the most important task for understanding documents and archives \cite{ref_lncs2}\cite{ref_article7}. In the context of historical document, furthermore, the name entity recognition is the most requested function by scholars searching for historical information \cite{ref_proc7}\cite{ref_article4}. However, NER for historical documents encounters more challenges. This is because historical documents often contain archaic language, obsolete terminologies, and variations in spelling and grammar that are not present in modern documents. Additionally, historical documents may contain named entities that are no longer in use or have changed in meaning over time, which can further complicate the NER process\cite{ref_article4}\cite{ref_proc9}\cite{ref_articles8}\cite{ref_proc11}\cite{ref_proc12}. Another challenge that makes NER more difficult for historical documents is the requirement for a large corpus of text data to. While modern languages are easily accessible through the internet, historical languages are not as readily available for searching and crawling. There are no comprehensive digital archives for historical languages, and the text data that does exist may be scattered across various physical and digital sources, making it difficult to compile a large corpus for NER\cite{ref_article4}\cite{ref_proc9}\cite{ref_article_labsch}. 

In this study, we introduce a Korean historical corpus called Seungjeongwon ilgi, the Royal Secretary's diary from the Joseon Dynasty, which was written in classical Chinese. This corpus could be useful for researchers or historians who are interested in studying the past and gaining insights into historical events and their context.
Utilizing the Seungjeongwon corpus for research purposes offers several important advantages. First, Seungjeongwon diary was written in classical Chinese. East Asia, including countries such as Vietnam, Japan, and Korea, classical Chinese characters were commonly used in official documents. The use of Chinese characters in official documents allowed for easier communication and record-keeping among East Asian countries. Consequently, the study's findings may have broader implications for the study of East Asian history and culture beyond the specific context of Seungjeongwon corpus. Second, Seungjeongwon diary is a record written by secretaries\ref{fig:teaser} who assisted the king in the Joseon Dynasty, which lasted for 500 years. As such, this corpus provides a unique and consistent perspective that allows for the observation of changes in vocabulary and expression over time from a diachronic point of view. Third, as Seungjeongwon diary is regarded representative historical record of the Joseon Dynasty, much of historians have worked to provide additional information about it. The document was originally written in a cursive style called wild grass, which has posed challenges for scholars seeking to decipher and analyze its content. To overcome these challenges, scholars have "decoded" the document by transcribing it into a more legible format, recording it in modern Unicode, identifying phrase identifying and organizing punctuation and important entity information contained within it. This variety of ancillary information can contribute to the interpretation of other uninterpreted historical information.

The structure of current study is as follows. (1) The second section is an introduction of the Seungjeongwon corpus, presenting statistical figures of corpus. (2) The third section presents a research model for the study, outlining the approach and methodology used to analyze the corpus. This section describes language model and context embedding language model, some of which is based on the FLAIR\cite{ref_article4}\cite{ref_proc5}library. (3) The fourth section involves testing the models under various conditions and analyzing and discussing the results of the tests. (4) Finally, the fifth section is the conclusion, where the findings of the study are summarized and their implications for future research are discussed.

\section{Corpus}
\subsection{History of Joseon Dynasty and SeungJeongWon Diary}
Joseon is a dynasty that ruled the Korean Peninsula from 1392 to 1897. It was founded by Yi Seong-gye, a general of the preceding Goryeo dynasty. Over the course of its history, Joseon was ruled by 26 different kings, and in 1897 it was renamed the Korean Empire, with the second emperor reigning until 1910 when the country was annexed by Japan. Seungjeongwon office was established at the beginning of the Joseon Dynasty and served as the Royal secretariat, handling all state secrets and sensitive administrative affairs. The Seungjeongwon diary was kept by the six representative secretaries of Seungjeongwon, recording royal orders, administrative affairs, and ceremonial matters that were handled during the Joseon Dynasty\cite{ref_url2}.

\begin{figure}
	\includegraphics[width=\linewidth]{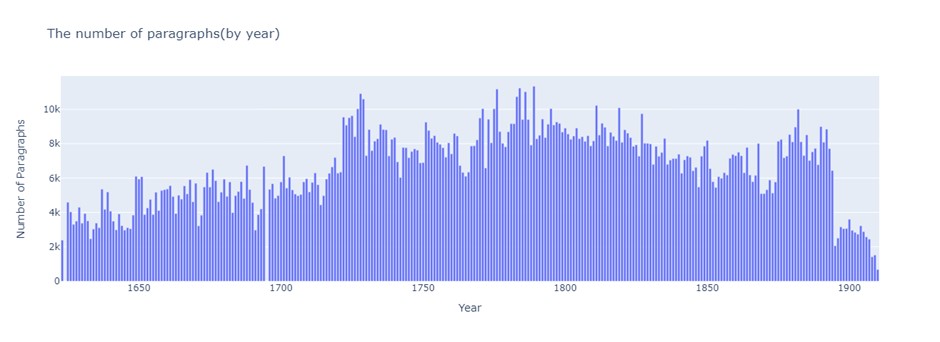}
	\caption{The number of paragraphs of Royal secretary’s diary over year} \label{fig1}
\end{figure}

The Seungjeongwon diary is considered more valuable than the Annals of the Joseon Dynasty, which is listed as a World Memory Heritage, and is regarded as a valuable document with only one original copy\cite{ref_url2}. It was used as a basic data when compiling the Annals, and is recognized as the world's largest historical material. In 2001, it was registered as a UNESCO World Heritage Site, highlighting its significance\cite{ref_url3}. The diary was written in a diary style, with one book per month, and was organized from the beginning of the Joseon Dynasty. However, the first part of the diary was lost due to war and other factors, and only 3,243 books remain from 1623 (Injo 1) to 1910 (Soonjong 4)\cite{ref_url1}.  The Fig.~\ref{fig1}. below shows the total number of paragraphs written between 1623 and 1910. 

The Seungjeongwon diary holds great historical value as it is the largest chronological record in the world, consisting of 3,243 books and 242.5 million characters. This is a much larger volume than China's 25 chronicles, which consist of 3,386 books and about 40 million characters\cite{ref_book1}, as well as the Veritable Records of the Joseon Dynasty, which comprise 888 books and 54 million characters\cite{ref_url2}). In specific, the Seungjeongwon Diary corpus contains a total of 1,896,173 paragraphs and includes 13,666 types of characters, excluding special characters. While most characters used are in classical Chinese, there are also around 250 Korean characters. The longest paragraph in the corpus consists of 36,992 characters, and on average, each column contains 118 characters\footnote{Seunjeonwon Diary corpus is freely available after registration from www.data.go.kr.}.

\subsection{Historian’s NER annotation and punctuation marker }

The Seungjeongwon diary was not widely recognized and utilized before the 1960s, despite its high value. The main reason for this was the difficulty in reading and understanding the content, as it is the only copy and written in wild cursive Chinese (Fig.~\ref{fig:teaser}). Only a few Chinese scholars who are well-versed in cursive writing were able to comprehend its contents\cite{ref_article6}.

The National History Compilation Committee undertook the task of decoding the cursive writing and published 141 English copies of the Seungjeongwon Diary through the Seungjeongwon diary Re-publishment Project between 1960 and 1977. They also began to digitize the original Seungjeongwon Diary and make it available on the web . Therefore, for historical research purposes, there are two versions of the data: the original cursive-style document and the reinterpreted clearly copied version in a more readable format\cite{ref_article6}.

In order to make the Seungjeongwon Diary more accessible for historical research, the clearly copied document were digitized into a computer-readable format using Unicode. Additionally, starting in 1993, the Korean Classical Translation Institute began translating the diary. However, due to its vastness and complexity, the complete translation of this historical material has yet to be finished even as of 2023, 30 years after the translation began\cite{ref_article6}. The original documents contain no punctuation marks except for space ahead of special character of “king” in Chinese(Fig.~\ref{fig2}). The corpora were manually annotated by historian according to national guidelines, and historians have identified three types of objects, namely, name, place, and book title, and attached marker to distinguish the phrase and clarify meaning. There are 3 types of named entities which include name, location and title of book and there are at least five types of special markers.

Historians also have attached additional notes. Omissions notes mean letters that are filled in for purpose of completing sentence. Comparative notes refer to letters that have been corrected or filled in with correct content compared to other books. Linking notes refers to letters that are unnecessary but are judged to be included.

In sum, the Diary Records of Royal Secretariat of Joseon Dynasty corpus is provided by National Institute of Korean History. It contains records from 1623(Injo) to 1910(Soonjong), with a total of 3,243 books and 3,186 volumes. We used text and label information of named entities from the corpus. The text has 1,896,173 paragraphs and 13,666 characters. The average length of paragraphs is 118 characters. We download all of corpus and prepared two kinds of corpus, one which is with punctuation and the other is without punctuation marker. We only use two King’s diary, namely Injo and Soonjong. 

\begin{table}
	\caption{Descriptive statistics for each of Injo and Soonjong diary}\label{tab1}
	\begin{tabular}{cclllll}
		\toprule
		King (Period) &   Characters & Person & Location & Book \\
		\midrule
		Injo (1596 - 1649)  & 7,156,961 & 228,394 & 40,665 & 9,105\\
		Soonjong (1907 - 1910)  & 513,221 & 38,754 & 11,648 & 87\\
		\bottomrule
	\end{tabular}
\end{table}

\section{Research Design}
\subsection{Research Model}

We intend to examine what is the most efficient and suitable method for interpreting historical information by implementing and experimenting with NER that predicts the future corpus based on current language knowledge.

For this study, we borrowed the concept of the basic idea of transfer learning. Transfer learning is a powerful concept in machine learning where the knowledge and patterns learned from one task can be applied to another task. We extend the concepts from task to time perspectives. In specific, past document is used as a starting point to train NER model. By using the models trained by past document, the future document is deciphered. 

\begin{figure}
	\includegraphics[width=\linewidth]{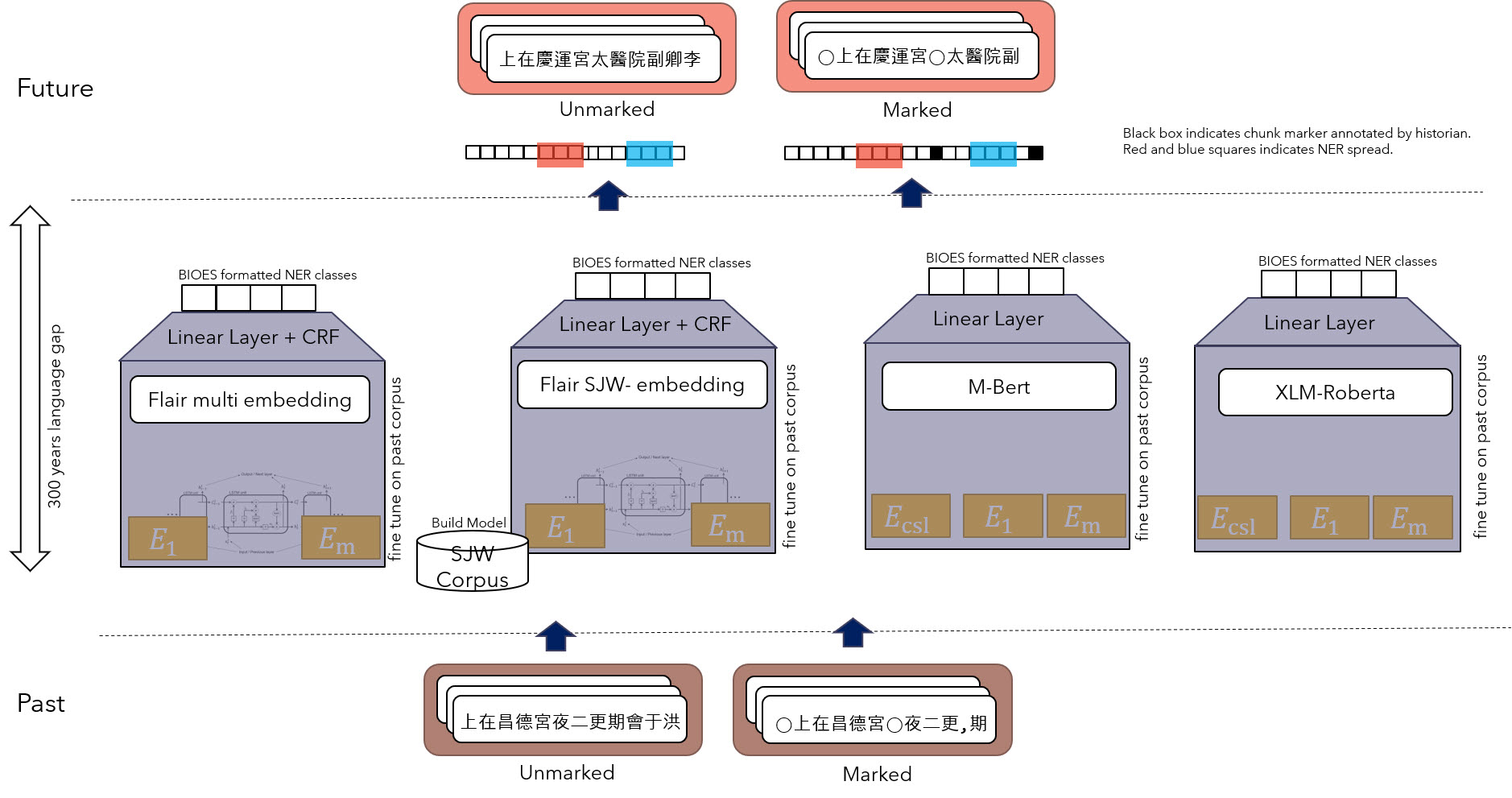}
	\caption{Process of transfer learning from past to future corpus} \label{fig3}
\end{figure}

We designed a research model investigating effective NER strategies, which is shown in Fig.~\ref{fig3}. The major threefold treatments are as follows. First, we apply a model trained on past data (bottom) to future data (top). The diary of Injo and Soonjong, which appeared in Tab.~\ref{tab1}, are a document of the past and the future respectively.  Second, we use four different pretraining models, namely (1) Flair multi-language embedding, (2) Flair Sengjeongwon-specific word embedding (here SJW-Flair), (3) muti-language BERT (4) and xml-ROBERTA. We also fine-tuned it on the past documents. Flair is a lightweight and flexible model that specializes in natural language processing (NLP), and one of its key strengths is its ability to stack various embedding layers including transformer based large model. We use also Flair library incorporate all of pretraining embedding model. Third, we prepare two difference style of corpus. The documents prepared for this study consist of two different styles: the original diary without punctuation marks and a version of the same diary with punctuation marks added by historians. 
In short, we test all of four models along with four kinds of corpus in combination of time and annotated marker.

\subsection{Hypothesis Setup}

\begin{figure}
	\includegraphics[width=\linewidth]{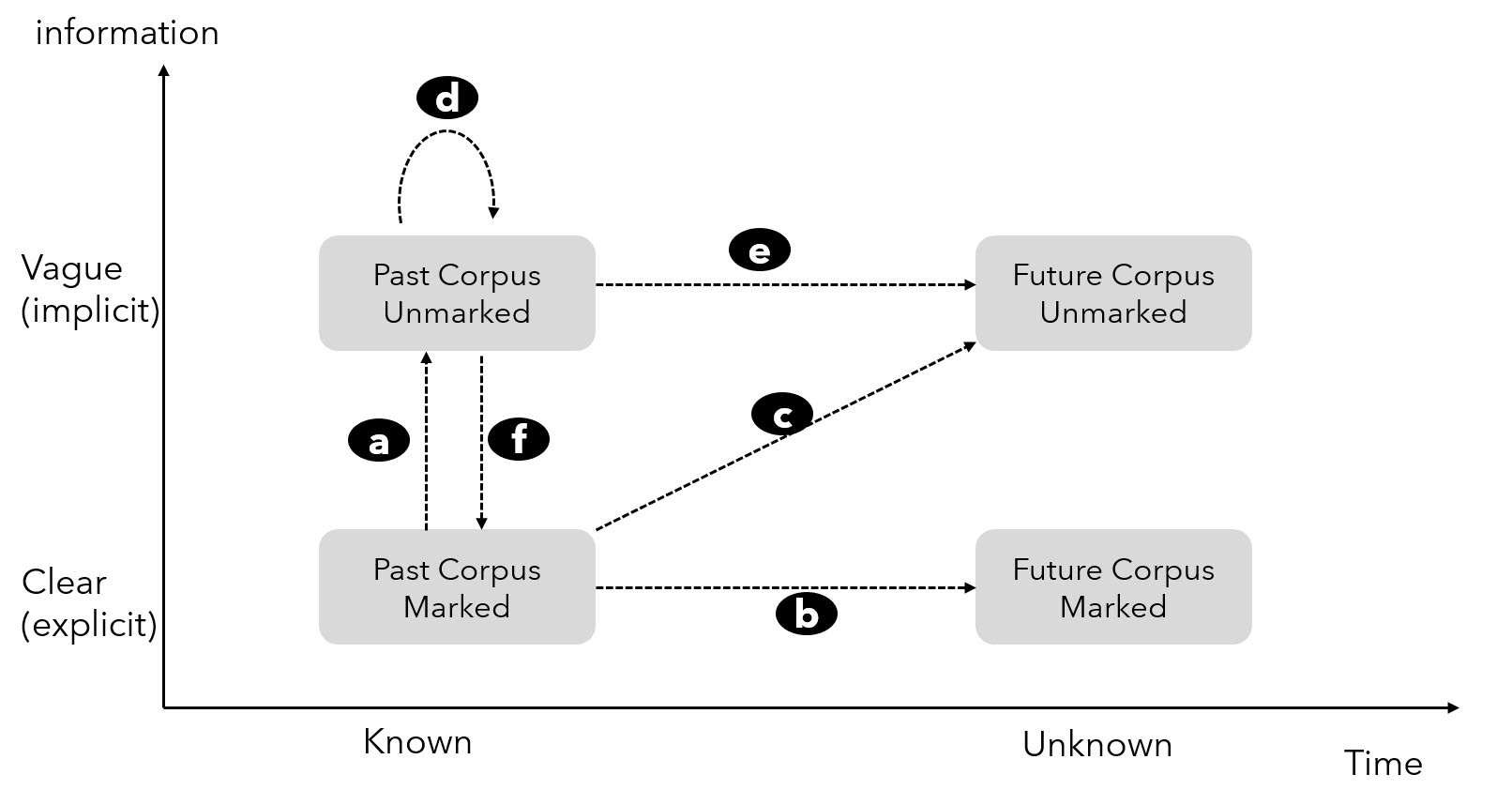}
	\caption{Define paths for comparison } \label{fig4}
\end{figure}

We argue that deciphering coarsely used corpus is much more difficult because of two major reasons which is related to two axes in Figure.~\ref{fig4}. 

One axis refers to “information”. It can be challenging to understand the semantic meaning without clear structure and relative information within context.   For example, if we don't even know where a sentence breaks, it will be much harder to interpret. Understanding the structure of a sentence is critical to interpreting its meaning accurately. This is especially true for historical language.  The presence of sentence boundaries and punctuation can significantly affect the interpretability of a corpus, especially in the context of historical texts. In general, a corpus with identified sentence boundaries and punctuation provides more definitive information, while a corpus with no identified sentence boundaries and no punctuation can provide relatively ambiguous information.

Another axis refers to “time”. Even if learned terms and usage are understandable at the present time, usage changes over time and becomes more difficult to interpret. We cannot easily understand conversations from just a hundred years ago. This is because language is not a static entity, but rather a dynamic system that evolves and adapts to changing social, cultural, and historical contexts. Linguistic patterns and named entities used in documents in same time period can provide information for understanding the context and meaning of the current documents. However, the usefulness of this information is limited to the time period in which the document was created.

Based on above mentioned perspective, we set up six transfer learning paths from (a) to (f) which is show in Fig.~\ref{fig4}. 

All of these paths split into two different training strategies. First strategy is based on punctuation marked corpus. A refers to training past/marked corpus condition and applying it to past/unmarked condition. B refers to applying same training model to future/marked condition. C refers to applying same training model to future/unmark condition. 

Second strategy is based on unmarked corpus. D refers to training past/unmarked corpus condition and applying it to past/unmarked condition. E refers to applying same training model to future/unmarked condition. F refers to applying same training model to past/marked condition. 

Given above mentioned perspectives and six paths, we set up four hypotheses as follows

\subsubsection{Hypothesis 1}{\bfseries D} is better than {\bfseries A} as annotated mark in corpus may be a noise in predicting unmarked corpus.

\subsubsection{Hypothesis 2} {\bfseries C} is better than {\bfseries E} as annotated corpus clarify a pattern to predict unknown future.

\subsubsection{Hypothesis 3} {\bfseries B} is better than {\bfseries E} as annotated corpus clarify a pattern to predict unknown future.

\subsubsection{Hypothesis 4} {\bfseries A} is better than {\bfseries F} as the model learned more patterns can predict the corpus less complicated. 

\subsection{Building Language Model and Fine Tuning on past corpus}

To create a language model specialized for Seungjeongwon Diary, four types of embedding were used and pre-trained with Injo's diary. Tab.~\ref{tab2} specifies the size of each model, the amount of time taken for fine-tuning, the loss function used and number of parameters.The four types of embedding used are as follows:

1.	M-Flair: This is a language model that generates contextualized word embeddings by considering the surrounding words in a sentence\cite{ref_articles8}. M-Flair is pre-trained on multilingual languages using JW300 Corpus\cite{ref_proc3}.

2.	SJW-Flair : This is a Flair language model but trained from only for Seugjoenwon Diary. We used the flair library to create an LSTM-based forward and backward language model.

3.	M-BERT: This is pre-trained on a large corpus of text data such as books and wikipedia written in various languages\cite{ref_article2}.

4.	XLM-ROBERTA : This is a pre-trained language model based on the XML(cross-lingual language model) and RoBERTa\cite{ref_article_liu}\cite{ref_article6}. It is pretrained by 2.5 terabytes of text data using common crawl.

\begin{table}
	\caption{Descriptive statistics for model comparison}\label{tab2}
	\begin{tabular}{ccccc}
		\toprule
		model & model size & train time & epoc/bs &  parameters\\
		\midrule
		M-Flair 	&	26M	&	2:13	&	10/32		&		6M		\\
		SJW-Flair 	& 264M 	&   3:47 	& 	10/32	 	& 	 	95M		\\
		M-BERT 		& 682M 	&	30:30	&	10/32	 	& 	 	177M	\\
		XLM-ROBERTA	&	1.1G&	35:10	&	10/32	 	& 	 	278M	\\
		
		\bottomrule
	\end{tabular}
\end{table}

\section{Results and Analysis}

NER performance was evaluated for each path and condition according to the embedding used. The micro F1 score was used as the primary metric, but the Flair library also provided accuracy and quality scores for each entity. It appears that the performance of the Transformer embedding was generally excellent, even with the addition of a CRF layer\cite{ref_article_huang}, when compared to the Flair embedding. However, in the case of the SengjeongwonFlair embedding, which utilized a corpus-specific language model, the performance was superior to the Flair fine-tuning approach. 

\begin{figure}
	\includegraphics[width=\linewidth]{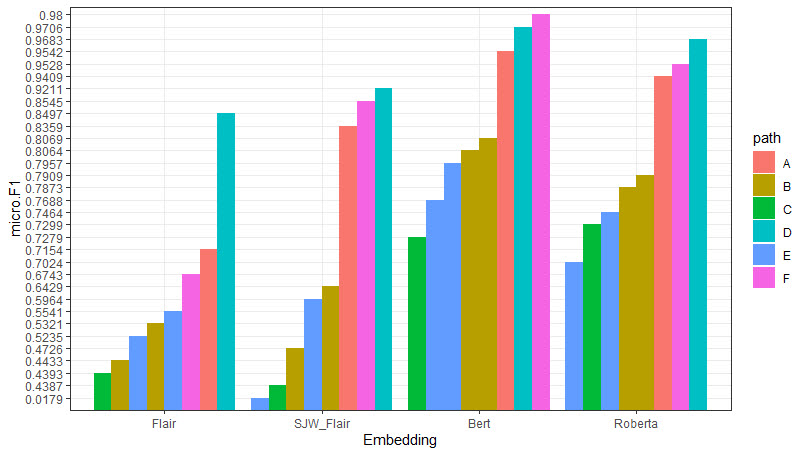}
	\caption{Micro F1 score across embedding and various paths} \label{fig5}
\end{figure}

Among the six transfer learning paths presented in Section 2.3, {\bfseries (d)} is the only path that predicts NER with same time period. Therefore, {\bfseries (d)} is predicted to be the best in all kinds of models. However, unlike all other models, the BERT model showed higher accuracy of the {\bfseries (f)} path than that of {\bfseries (d)} . That is, the model that learned the entity name based on the corpus to which the punctuation was assigned predicted the entity name more accurately than the model without the mark. The characteristic of BERT's pre-training algorithm is to predict masked letters. Due to this feature, BERT seems to learn more clearly in the marked state and even handles more clearly on unmarked data.

The comparison between two transformer-based models, BERT and XLM-Roberta, revealed that BERT outperformed XLM-Roberta in most of the pathways, except for pathway {\bfseries (c)} which intersects all of conditions of time and information. BERT was pre-trained using official documents like Wikipedia, whereas XLM-Roberta was trained by crawling data and is known to perform better on colloquial language. The results suggest that a language model trained on modern written language style can better fit even for past text content. However, it is worth noting that XLM-Roberta, which is designed to be more suitable for cross-lingual tasks, did not show significant performance deterioration in handling the exchange of information and time, as shown in Fig.~\ref{fig5}.

In comparing the two Flair model bases, M-Flair and SJW-Flair, it was found that SJW-Flair performs better for documents of the same era but worse for documents of different eras. Specifically, SJW-Flair showed higher performance than M-Flair in predicting past documents with past documents in the {\bfseries (a)}, {\bfseries (f)}, and {\bfseries (d)}  paths. This is likely due to the fact that SJW-Flair was trained exclusively on the Seungjeongwon Diary corpus, which is from the same era as the documents being predicted. However, for predicting past documents with modern documents in the {\bfseries (b)}, {\bfseries (c)}, and {\bfseries (e)}  paths, M-Flair outperformed SJW-Flair. This is because M-Flair was pre-trained on a larger and more diverse corpus, including modern language, which allowed it to better handle documents from different eras.
Further analysis on each hypothesis is conducted on the preceding chapter\footnote{Authors supply the best History NER model, which is Bert Path A at https://huggingface.co/Nara-Lab/History\_NER}  

\subsection{Hypothesis H1 (supported)}

This experiment compares models that recognize named entities in documents without punctuation and models that recognize named entities in documents with punctuation. The goal is to see if adding unnecessary information in the training data harms the model's predictive power. The two boxplots below show quality scores by embedding type and pathway. The quality score includes all quality metrics provided by Flair in addition to Micro F1 score.

\begin{figure}
	\includegraphics[width=\linewidth]{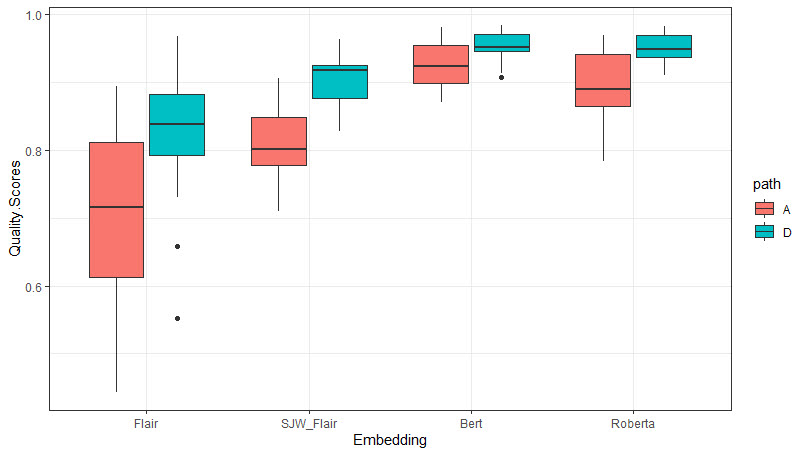}
	\caption{Quality scores over embedding(H1: A and D)} \label{fig6}
\end{figure}

When comparing {\bfseries (a)} and {\bfseries (d)} in the second figure that compares models, {\bfseries (d)}'s scores are generally higher. Although BERT, which is characterized by masking character recognition, has the feature that the performance of the pathway predicting documents without punctuation from documents with punctuation is not significantly reduced, the overall performance of the {\bfseries (d)} pathway is good as shown in Fig.~\ref{fig6}.

\subsection{Hypothesis H2 (not supported)}

\begin{figure}
	\includegraphics[width=\linewidth]{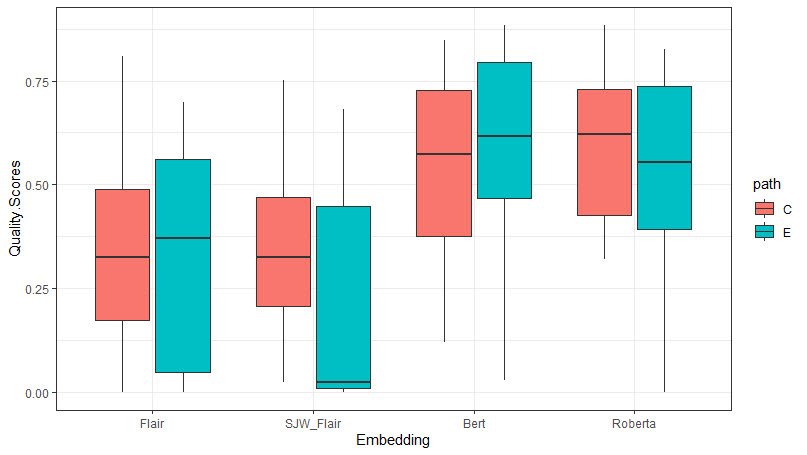}
	\caption{Quality scores over embedding(H2: C and E)} \label{fig7}
\end{figure}

The experiment aims to extend the previous experiment in the temporal direction. It verifies whether there is a difference in performance depending on the pres ence or absence of punctuation when applying a model trained on past corpora to future corpora. In Fig.~\ref{fig7}, there is no significant difference between the two values when comparing {\bfseries (c)} and {\bfseries (e)}. In the second figure, it is observed that the performance of path {\bfseries (e)}, which predicts documents without punctuation from documents with punctuation, is slightly higher for BERT, which has the feature of masking character recognition, but no significant difference is observed overall.

\subsection{Hypothesis H3 (supported)}

\begin{figure}
	\includegraphics[width=\linewidth]{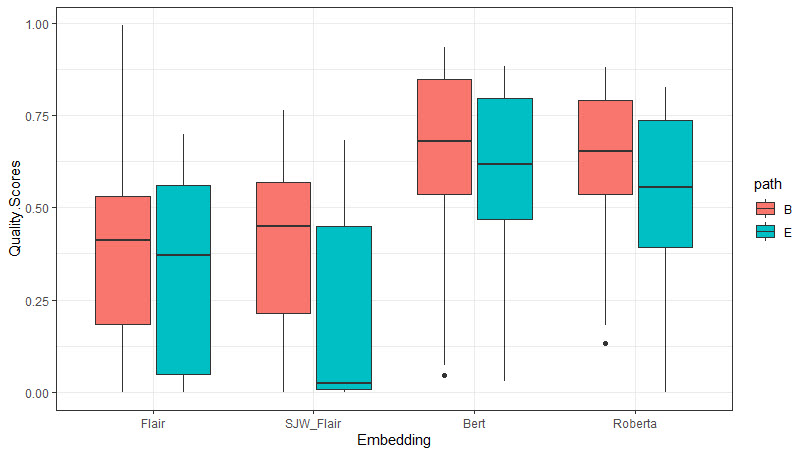}
	\caption{Quality scores over embedding(H4: B and E)} \label{fig8}
\end{figure}

This experiment evaluates the performance of models with and without punctuation when predicting future corpora based on past corpora. Unlike H2, which evaluated predicting future corpora from past corpora, there is no cross-style of document writing. In contrast to the previous experiment, H3 consistently shows that the model with punctuation performs better (See Fig.~\ref{fig8}).

\subsection{Hypothesis H4 (not supported)}

The purpose of this experiment is to evaluate the performance difference between two models. One is trained on punctuated documents and applied to not-punctuated documents and the other is trained on not-punctuated documents and applied to punctuated documents. 

\begin{figure}
	\includegraphics[width=\linewidth]{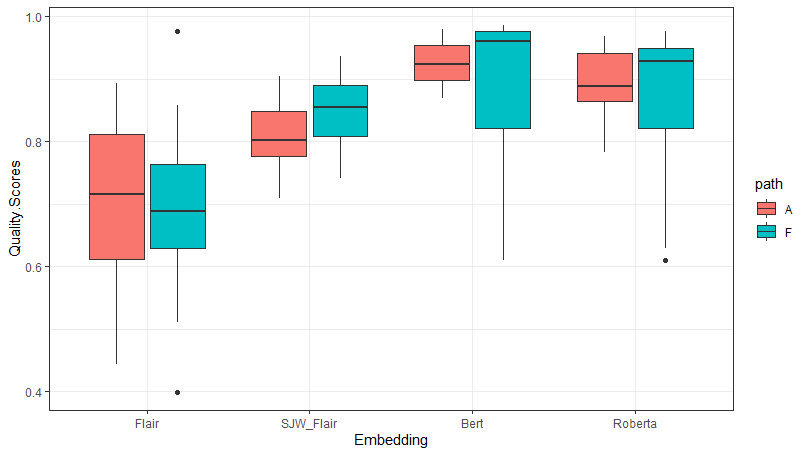}
	\caption{Quality scores over embedding(H4: A and F)} \label{fig9}
\end{figure}

As shown in the Fig.\ref{fig9}, except for M-FLAIR, all models tend to perform better when they use more information and are pre-trained, and then predict less informative documents, as opposed to the case of {\bfseries (a)} where they predict more informative documents with less information.

\subsection{Summary of Results }
In this chapter, the final results of the t-test are described using the quality values for each condition examined above. The data used for the welch’s t test\cite{ref_articles9} using all of quality numerical evaluation scores provided by Flair\cite{ref_articles8}. Information summarizing the test results for hypotheses follows. First, when trying to recognize an entity name in a corpus without punctuation marks, the performance of the model trained with the same corpus without punctuation marks is good ({p < 0.005}). Second, when extracting entities for non-contemporaneous data, scoring all data performed better than not assigning punctuation marks (p < 0.005)
The summarized results are shown in the following table:

\begin{table}
	\caption{Results of Hypothesis Test}\label{tab3}
	\begin{tabular}{cccccl}
		\toprule
		Hypothesis & t.stat & df & p-value &  Mean & Results\\
		\midrule
		H1	& 	-4.2575	&	119.27	&	0.0000	&		A:0.8338	 & Supported\\
		&			&			&			&			D:0.9055	&	\\

		H2	& 	0.9614	&	155.62	&	0.3379	&	C:0.4556	 & not Supported\\
		&			&			&			&			E:0.4182	&	\\
		
		H3	& 	2.8472	&	267.34	&	0.0048	&	B:0.5141	 & Supported\\
		&			&			&			&			E:0.4182	&	\\
		
		H4	& 	0.4876	&	128.49	&	0.6268	&		A:0.8339	 & not Supported\\
		&			&			&			&			F:0.8233	&	\\
	
		\bottomrule
	\end{tabular}
\end{table}

\section{Conclusion}
Our study introduced the historical corpus which is annotated with NER as well as punctuation marker. Our study also conducted experiments on applying transformer models and explored the effectiveness of various embedding techniques to predict named entity. Furthermore, we developed the research model to investigate whether the learned knowledge from past context transfer to future contexts. Our results shows that the punctuation information provided by historian annotators had improve the performance of NER. In our comparative study, we evaluated the performance of different NER strategies using the Flair library and we used a classic statistical test known as Welch's t-test to determine the significance of the differences in performance between the models. We believe that our study will serve as a meaningful piece of research to tackle the challenges of historical data.  

In the future study, it will be necessary to examine in more detail which types of entities are better identified in the context of transfer learning across far or close time gap. Our research findings also suggest other type of strategies holding interaction between historian and machine might be adventurous. Given the scarcity of historical corpus, authors believe various practical tactic to handle such scares corpus should be researched in the future.

\bibliographystyle{ACM-Reference-Format}
\bibliography{history_NERbb}


\section{Online Resources}

Resources are available here

Seungjeongwon Corpus : https://sjw.history.go.kr

Seungjeongwon NER : https://huggingface.co/Nara-Lab/History\_NER

\end{document}